\documentclass[letterpaper]{article} 
\usepackage{aaai24}  
\usepackage{times}  
\usepackage{helvet}  
\usepackage{courier}  
\usepackage[hyphens]{url}  
\usepackage{graphicx} 
\usepackage{natbib}  
\usepackage{caption} 
\usepackage{algorithm}
\usepackage{algorithmic}

\usepackage{newfloat}
\usepackage{listings}
\DeclareCaptionStyle{ruled}{labelfont=normalfont,labelsep=colon,strut=off} 
\floatstyle{ruled}
\newfloat{listing}{tb}{lst}{}
\floatname{listing}{Listing}
\title{Adventures of Trustworthy Vision-Language Models: A Survey}
\author{
    Mayank Vatsa,
    Anubhooti Jain,
    Richa Singh
}
\affiliations{
    IIT Jodhpur, India\\

   mvatsa@iitj.ac.in, jain.44@iitj.ac.in, richa@iitj.ac.in
}

\usepackage{bibentry}

\begin{document}

\maketitle

\begin{abstract}
Recently, transformers have become incredibly popular in computer vision and vision-language tasks. This notable rise in their usage can be primarily attributed to the capabilities offered by attention mechanisms and the outstanding ability of transformers to adapt and apply themselves to a variety of tasks and domains. Their versatility and state-of-the-art performance have established them as indispensable tools for a wide array of applications. However, in the constantly changing landscape of machine learning, the assurance of the trustworthiness of transformers holds utmost importance. This paper conducts a thorough examination of vision-language transformers, employing three fundamental principles of responsible AI: Bias, Robustness, and Interpretability. The primary objective of this paper is to delve into the intricacies and complexities associated with the practical use of transformers, with the overarching goal of advancing our comprehension of how to enhance their reliability and accountability. 

\end{abstract}

\section{Introduction}

\begin{figure}[t]
    \centering
    \includegraphics[width=1\linewidth]{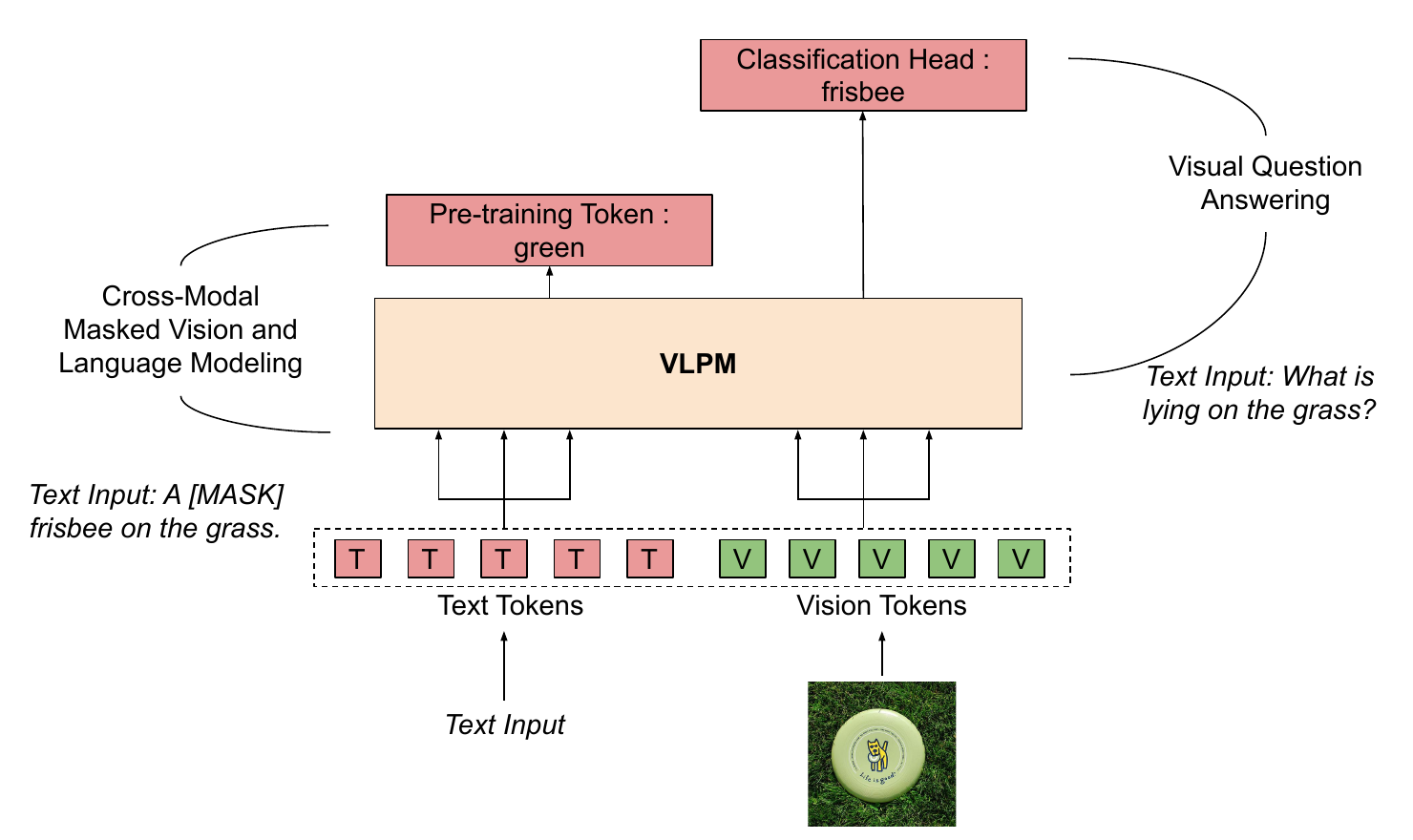}
    \caption{An example of vision-language model pre-trained using Cross-Modal Vision-Language Modeling and finetuned for Visual Question Answering.}
    \label{fig:barplot1}
\end{figure}

\begin{figure}[]
    \centering
    \includegraphics[width=0.97\linewidth]{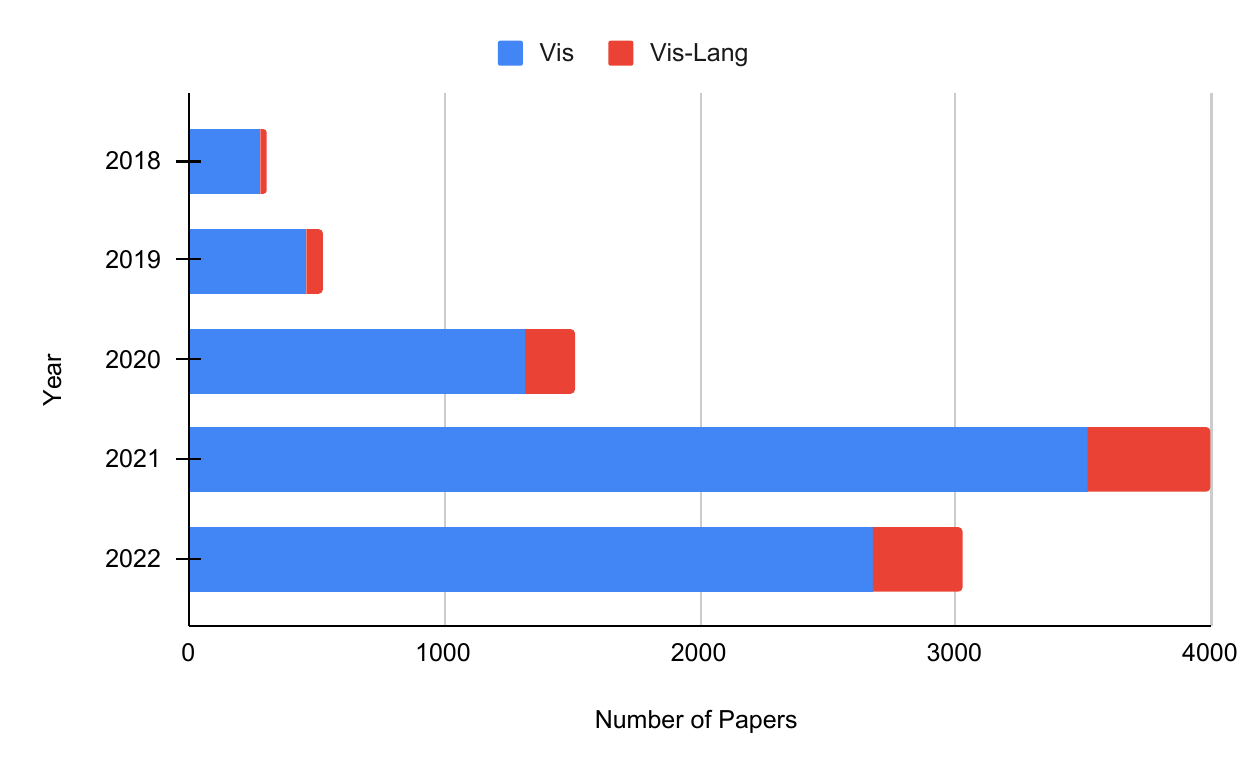}
    \caption{Keyword Analysis for research papers pertaining to two keywords, `vision-language transformer' (red) and `vision transformer' (blue) from 2018 to 2022.}
    \label{fig:barplot}
\end{figure}

Inspired from the performance for language-based tasks \cite{DBLP:conf/nips/VaswaniSPUJGKP17, DBLP:conf/naacl/DevlinCLT19}, transformers were proposed for vision-based tasks where they process images as patch tokens \cite{DBLP:conf/iclr/DosovitskiyB0WZ21}. 
Even with the modality change the basic architecture remained the same. These architectures were further extended to accommodate both modalities, giving birth to transformer-based vision-language models (Figure \ref{fig:barplot1}). Their self-attention module makes convolutions unnecessary, with \cite{park2022how} stating that multi-head self-attention acts as low-pass filters while convolutions act like high-pass filters. Their impressive success has been attributed to their ability to model long-range dependencies and having weak inductive biases, leading to better generalization. 
\cite{ijcai2022p773} discusses a general architecture for the Vision-Language Pre-trained Models (VLPMs), breaking the architecture into four categories, namely, Vision-Language Raw Input Data, Vision-Language Representation, Vision-Language Interaction Model, and Vision-Language Representation. \cite{ijcai2022p773,ijcai2022p762,DBLP:journals/corr/abs-2307-03254} surveys VLPMs based on their architecture, pre-training tasks and objectives, and downstream tasks, showcasing that VLPMs continue to grow not only in terms of accuracy but size as well, as the newer models have parameters in billions and can perform several tasks with human-like accuracy. As shown in Figure \ref{fig:barplot}, compared to 2018, there has been a big surge in articles about ``vision-language transformer'' in 2022, nearly 9.5 times more, and an even larger increase, nearly 12.5 times more, in 2021. A similar trend is seen with the term `vision transformer,' with roughly 15 times more articles in 2022 compared to 2018 and an astounding approximately 21 times more in 2021. Many of these models are trained on heavy open-web datasets and are finetuned for different tasks ranging from classification-based to generative-based.

\cite{ross2020measuring,birhane2021multimodal,srinivasan-bisk-2022-worst} have shown that these heavy and high-performing models suffer from different biases like gender and cultural bias. A detailed review of one of the vision-language transformers by \cite{srinivasan-bisk-2022-worst} depicts gender bias, with purse being the preferred term for the female gender while briefcase being the preferred term for the male gender. Just like bias, cases can be made for robustness and interpretability, iterating a need for a proper study of transformer models. 
Efforts have been made to study transformers in this light for vision and language-based models individually, but collectively, there are only a few studies so far. Hence, we present an extensive survey of these VLPMs from a dependability and trust point-of-view by curating different practices, methods, and models proposed for VLPMs, first expanding on bias, followed by robustness, and finally, interpretability. In the end, we also discuss open challenges in the field. With this study, we hope to present the current state of VLPMs regarding reliability and highlight some research gaps that can help alleviate the overall state of VLPMs. 


\subsection{An Overview of VLPMs}

In VLPMs, both single and dual architecture models have emerged as powerful tools. Here, we present a brief overview of these architectures and various pre-training and downstream tasks.

\begin{figure}[t]
    \centering
    \includegraphics[width=1.02\linewidth]{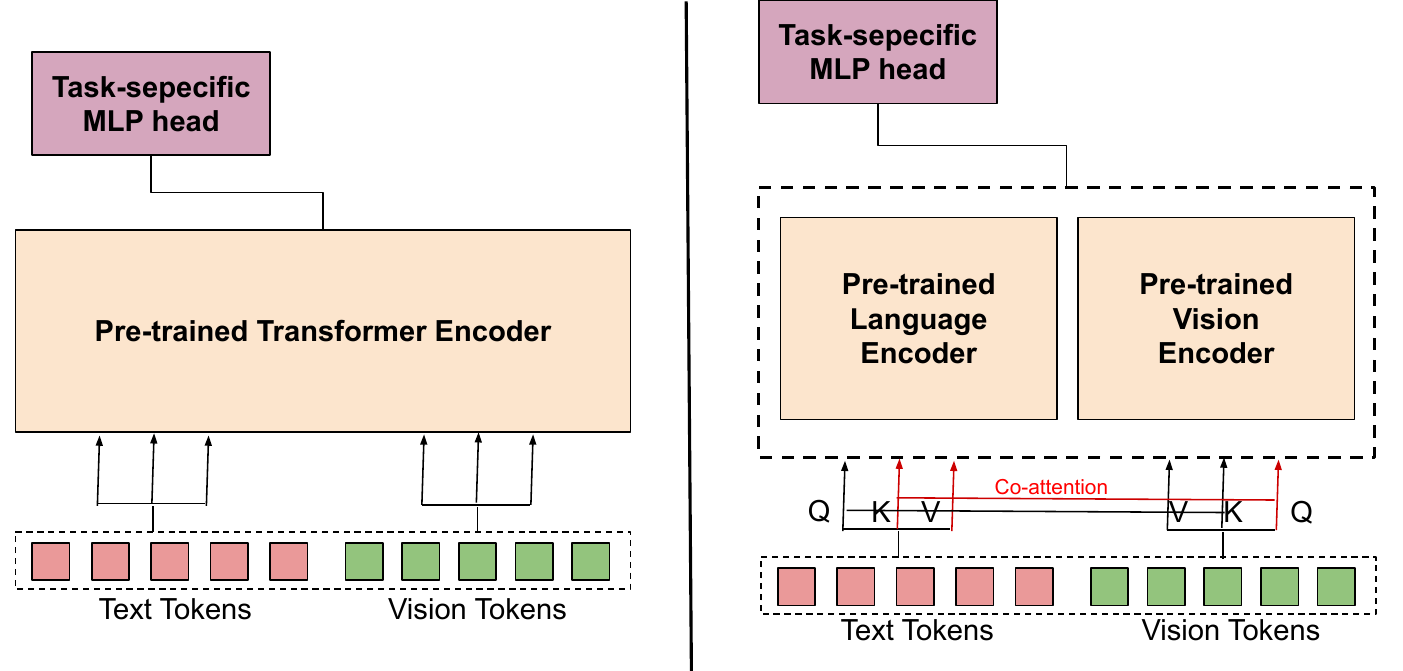}
    \caption{Generic single and dual-stream architecture for pre-trained vision-language transformer Models, The tokens represented in the figure are after including the positional embeddings.}
    \label{fig:vislang}
\end{figure}

\paragraph{Single and Dual Architectures:} While VLPMs have their own different architectures, they can be broadly categorized into two types of architectures (Figure~\ref{fig:vislang}). Single-stream models fuse both modalities early on with a single transformer using joint cross-modality like VisualBERT \cite{li2019visualbert} and ViLT \cite{kim2021vilt} transformer models. Dual-stream models, on the other hand, process the two modalities separately and are then modeled jointly, like ViLBERT \cite{lu2019vilbert} and LXMERT \cite{tan2019lxmert} models. VLPMs can also be divided on the basis of visual features extracted from the model, like region features, usually pulled from object detectors, used by models like ViLBERT~\cite{lu2019vilbert}, grid features used by models like Pixel-BERT~\cite{huang2020pixel}, or patch projection used by models like ViLT~\cite{kim2021vilt}. 

\paragraph{Pre-training Tasks:} Pre-training has been found to be very beneficial for transformers and, by extension, for VLPMs. The models are pre-trained on large datasets to solve different pre-training tasks in a supervised or self-supervised fashion. VLPMs generally use image-caption pairs for pre-training using paired as well as unpaired open web datasets, depending on the pre-training task. One of the most common tasks used for pre-training in the language models is Cross-Modal Masked Language Modeling, and it can be easily mapped for cross-modality in the vision-language domain as well. The task is generally used in a self-supervised setting where some tokens are masked randomly, and the goal is to predict the masked tokens. Another common task is Cross-Modal Masked Region Modeling, where tokens are masked out in the visual sequence. Cross-modal alignment is a task where the goal is to try to pair image and text, also known as Image-Text Matching (ITM). Cross-modal contrastive Learning is another pre-training task quite similar to ITM but in a contrastive manner in the way that matched image-text pairs are pushed together and non-matched pairs are pushed apart using contrastive loss. The large datasets used for pre-training have been considered to be a cause of bias \cite{park2022explanation,radford2021learning}.

\begin{table*}[!t]
\caption{Summarizing research studies that have proposed different bias metrics.}
\begin{tabular}{|l|l|l|}
\hline
\textbf{Bias Study}                                                                                    & \textbf{Models Under Review}                       & \textbf{Bias Metric}                                                                                      \\ \hline
\begin{tabular}[c]{@{}l@{}}Social Bias (Gender and Race) \\ \cite{ross2020measuring}    \end{tabular}                                                                         & ViLBERT, VisualBERT                                & Grounded SEAT and WEAT                                                                                    \\ \hline

Gender Bias \cite{DBLP:journals/corr/abs-2104-08666}                                                                                           & VLBERT                                             & Association Scores                                                                                        \\ \hline

\begin{tabular}[c]{@{}l@{}}Social Bias (Gender and Race) \\ \cite{DBLP:conf/cvpr/HirotaNG22}  \end{tabular}                                                                      & \begin{tabular}[c]{@{}l@{}} NIC, SAT, Att2In, UpDn, \\ Transformer, Oscar, NIC+ \end{tabular} & Leakage in Image Captioning (LIC)                                                                         \\ \hline

Social Bias \cite{zhang2022counterfactually}                                                                                            & ALBEF, TCL, ViLT                                   & CounterBias                                                                                               \\ \hline
\begin{tabular}[c]{@{}l@{}}Stereotypical Bias (Gender, Profession, Race, \\and Religion) \cite{zhou2022vlstereoset}\end{tabular} & \begin{tabular}[c]{@{}l@{}} VisualBERT, LXMERT, ViLT, \\ CLIP,  ALBEF, FLAVA  \end{tabular}    & \begin{tabular}[c]{@{}l@{}}vision-language relevance score and \\ vision-language bias score\end{tabular} \\ \hline
\begin{tabular}[c]{@{}l@{}}Quantifying bias before and after \\ finetuning \cite{DBLP:journals/corr/abs-2303-07615} \end{tabular}                & ResNet, BiT, CLIP, MoCo, SimCLR                    & Bias Transfer Score (BTS)                                                                                 \\ \hline
Emotional and Racial Bias \cite{DBLP:journals/corr/abs-2304-04874}                                                                              & \begin{tabular}[c]{@{}l@{}} NIC, SAT, Att2In, UpDn, \\ Transformer, Oscar, NIC+ \end{tabular}   & $ImageCaptioner^2$    \\ \hline                                                                     
\end{tabular}
\label{tab-bias-studies}
\end{table*}

\paragraph{Downstream Tasks:} Once VLPMs are pre-trained, they are finetuned to perform specific downstream tasks such as Image Captioning, Visual Question Answering, Image Text Retrieval, Natural Language for Visual Reasoning, and Visual Commonsense Reasoning. Broadly, the tasks can be categorized as generative, classification, and retrieval tasks. Task-specific datasets are used for finetuning the model, where the heads of the VLPMs are modified based on the downstream task. VLPMs have shown impressive accuracy with these tasks. The learned representation helps finetune the model for specified tasks quickly, especially with the rich information flowing between the two modalities.

We can draw two important observations from this overview of VLPMs: 

\begin{itemize}
    \item The architecture of VLPMs differs significantly from CNNs. Consequently, it's crucial to develop methods specifically tailored to the VLPM architecture rather than merely extending approaches originally designed for CNNs. This ensures a more accurate and equitable evaluation of their performance.
    \item  Most recent VLPMs undergo training on datasets derived from the open web, which is a combination of various sources. This amalgamation raises concerns about the potential incorporation of biases present in the content from the open web into the models themselves \cite{mittal_RMLD}.
\end{itemize}



\section{Bias and Fairness} 
Fairness in AI systems has been primarily viewed as protecting sensitive attributes in a way that no group faces disadvantage or biased decision. Biases like gender or racial bias have proven harmful, especially when they affect humans in real life \cite{bias_2022RS}. VLPMs are as vulnerable to bias as their CNN counterparts. They deal with two modalities and often two-stage training, allowing them to introduce more biases like pre-training bias or bias against a particular modality. Literature has shown that VLPMs are heavily influenced by language modality and can sometimes be harmful. \cite{kervadec2021roses} showed this with reference to the Visual Question Answering (VQA) task.

\subsection{Data and Bias} Data has been considered the primary source of bias as it is a representation of the world that the model is trying to learn. With VLPMs, this can be an even bigger issue as pre-training requires large datasets. Many well-known VLPMs today have been trained on large heavy datasets crawled from the Internet, giving less control and oversight during data collection. This can lead the dataset to learn harmful representations. \cite{DBLP:conf/iccv/ZhaoWR21} examines some widely used multimodal datasets for bias and shows offensive texts and stereotypes embedded within them. \cite{DBLP:journals/corr/abs-1912-00578} specifically examines dataset bias by studying the COCO dataset \cite{lin2014microsoft}, a manually annotated dataset for the image captioning task. The authors not only depict gender and racial bias but also analyze recent captioning models to see the differences in the performance from a lens of bias. Some studies have looked at task-specific datasets as well, as \cite{DBLP:conf/fat/HirotaNG22} analyze five Visual Question Answering (VQA) datasets for gender and racial bias. \cite{DBLP:conf/cvpr/GarciaHWN23} focuses on datasets crawled from the Internet without much oversight from a demographic point-of-view while also showcasing how societal bias is an issue on various tasks and datasets.  

\subsection{Bias Estimation and Mitigation}
\cite{sudhakar2021mitigating} studies biases present in vision transformers by visualizing self-attention modules, noting encoded bias in the query matrix. To study and mitigate these biases, they further proposed an alignment approach called TADeT. \cite{ross2020measuring} further measured social biases in the joint embeddings by proposing Grounded WEAT and SEAT while also proposing a new dataset for testing biases in the grounded setting. The study concludes that bias comes from the language modality, and vision modality does not help mitigate biases. Moreover, CLIP \cite{radford2021learning}, a heavily used VLPM known for its zero-shot capabilities, conducted its own bias study, postulating that it may encode social biases owing to the large open dataset used for its training. The authors tested zero-shot and linear probe instances of the model to mark the potential sources of biases and harmful markers. \cite{zhang2022counterfactually} proposes the CounterBias method and FairVLP framework to quantify social bias in VLPMs in a counterfactual manner while proposing a new dataset to measure gender bias. \cite{srinivasan-bisk-2022-worst} studies gender bias, particularly in the VL-BERT model, by modifying both language and vision modalities and getting association scores. They further create templates for entities to measure the bias in three instances - pre-training, visual context at inferencing, and language context at inferencing. It is particularly interesting as investigating the bias at different stages can not only help dissect the effectiveness of different modalities but can also allow examination of how VLPMs can evolve after the modalities integrate, giving a new perspective on merging the multiple modalities effectively. 


\cite{DBLP:conf/cvpr/HirotaNG22} introduced a new metric, Leakage for Image Captioning (LIC), to measure bias towards a particular attribute for the task of image captioning. The metric requires annotations for the protected attribute and can also use embeddings that have pre-existing bias. Furthermore, VLStereoSet \cite{zhou2022vlstereoset} measured stereotypical biases in VLPMs using probing tasks by testing their tendency to recognize stereotypical statements for anti-stereotypical images. The stereotype is based on four categories: gender, profession, race, and religion, making the VLPMs select the statements as captions. They also proposed two metrics called vision-language relevance score and vision-language bias score, using which they concluded that state-of-the-art VLPMs under consideration not only encode stereotypical bias but are more complex than language bias and need to be studied. Several studies have given mitigation techniques to deal with bias like \cite{DBLP:conf/eccv/HendricksBSDR18,DBLP:conf/bigdataconf/AmendWS21,DBLP:conf/icml/ZhaoAX23,DBLP:journals/corr/abs-2303-06167}. As can be noticed in these studies, there are different components and parts of the entire vision-language processing pipeline that are put under consideration. Even when looking for societal biases – gender and racial, there is a lack of commonality, yet none of the observations and results can be denied as less crucial. We feel that there is a lack of standard metrics and common protocol in the bias for multimodal models so far. In Table \ref{tab-bias-studies}, we have tried to summarize some of these studies, detailing the metrics they used and the models they examined for bias. VLPMs can encode bias with more opportunities to do so than unimodal models.


\section{Robustness} 
While accuracy focuses on correctness, robustness focuses on security by assessing the model for vulnerabilities in adversarial settings \cite{biasattack_2020RS}. Like CNNs, transformers are vulnerable to adversarial attacks. We first discuss how transformers perform against their CNN counterparts. Many have formulated that transformers are more robust than CNNs, but we believe that architectural differences were not considered by the adversarial methods used for these studies. We discuss the robustness of VLPMs exclusively in a separate subsection. 

\subsection{Transformers vs CNNs} 
Several transformer architectures have performed better than CNNs, \textit{but are they more robust?} \cite{bhojanapalli2021understanding} measures the robustness of ViT architectures to answer this very question and compares them with their ResNet counterparts for the task of image classification. Perturbations are added to the input using adversarial settings to measure robustness. The robustness is measured in parts, starting with natural perturbations like blurring, digitizing, and adding Gaussian noise. It is then measured with respect to distribution shift and using adversarial attacks. All the comparisons are made across varying sizes of ViT and ResNet architecture, concluding that transformers have a slight edge compared to ResNets, and with sufficient data, VITs can outperform their ResNet counterparts. \cite{shao2022on} studied the robustness of transformers by exposing them to white-box and transfer adversarial attacks, concluding that ViTs are more robust than CNNs. The study also observes that VITs have spurious correlations and are less sensitive to high-frequency perturbations. Adding tokens for learning high-frequency patterns in ViTs improves classification accuracy but reduces the robustness of the architecture. 

Hybrid architectures combining ViTs and CNNs can reduce the robustness gap between the two architectures. Most of the studies focus on transfer attacks in lieu of specific attacks for transformers. \cite{bai2021transformers,DBLP:conf/eccv/PintoTD22} studies the robustness between transformers and CNNs questioning previous studies \cite{bhojanapalli2021understanding,shao2022on} that show transformers to be more robust than CNNs claiming unfair settings while comparing the architectures. The study shows that transformers are not more robust than CNNs, but on out-of-distribution samples, transformers outperform CNNs. \cite{mao2022towards} proposed a Robust Vision Transformer (RVT) after studying the components affecting the robustness of the model, proposing a new patch-wise augmentation and a position-aware attention scaling (PAAS) to boost the RVT other than modifying damaging elements in the architecture for better robustness. RVT can be used as a backbone or vision encoder for different VLPMs, just like the Trade-off between Robustness and Accuracy of Vision Transformers (TORA-ViTs) \cite{DBLP:conf/cvpr/LiX23} that can combine predictive and robust features in a trade-off manner. \cite{mishra2022pretrained} performed a comparative study to measure the robustness of pre-trained transformers on noisy data. The noisy data is created using poison attacks like label flipping and has been compared under adversarial filtering augmentation. They introduced a novel robustness metric called Mean Rate of change of Accuracy with change in Poisoning (MRAP), using which they observed that the models are not robust under adversarial filtering. In most of these studies, the comparison between CNNs and transformers is drawn from existing attacks proposed originally for CNNs, but it is important to devise attacks that exploit vulnerabilities of the latter, keeping in mind the critical architecture difference between the two.

\subsection{VLPMs and their Robustness}  
VLPMs are studied under the robustness lens but not as extensively as unimodal transformers. \cite{li2020closer} studies VLPMs over linguistic variation, logical reasoning, visual content manipulation, and answer distribution shift. These models have already shown better performance in terms of accuracy. Still, for robustness, the authors propose an adversarial training strategy called MANGO or Multimodal Adversarial Noise Generator to fool the models. Further, efforts have been made to devise methods exclusively for transformers, like the Patch-wise Adversarial Removal (PAR) method \cite{shi2021decision} that processes each patch separately to generate adversarial samples in a black-box setting. The patches are processed based on noise sensitivity and can be extended to CNNs as well. \cite{ DBLP:conf/iccv/LiLG021} proposed a new benchmark for adversarial robustness on the task of VQA. \cite{wei2022towards} proposed a dual attack framework, namely, the Pay No Attention (PNA) method and PatchOut Attack, to improve the transferability across transformers that skipped attention gradients in order to create adversarial samples. Since the attack framework is sensitive to the transformer architecture, the attacks consider both patches by perturbing only a subset of them at each iteration and attention module by skipping some attention gradients. 

Other than attacks, \cite{ma2022multimodal} investigated how VLPMs perform under data with missing or incomplete modalities (examining only one modality at a time) in terms of accuracy and were improved using different fusion strategies. They concluded that transformers are not only sensitive to missing modalities but also that there is no optimal fusion strategy as multimodal fusion affects the robustness of these models and is dependent on datasets. \cite{DBLP:conf/aaai/SalinFAF22} analyzes VLPMs to get a better insight into the multimodal relationship using probing tasks, concluding that concepts like position and size are difficult for the models under consideration to understand. \cite{ DBLP:journals/corr/abs-2305-16934} studies adversarial vulnerability in a black-box setting to perform a realistic adversarial study by manipulating visual inputs. \cite{DBLP:journals/corr/abs-2308-10741} on the other hand, studied adversarial robustness for imperceivable attacks on VQA and Image captioning tasks for well-known multimodal foundation models and \cite{ DBLP:conf/iclr/MaoGYWV23} studies the zero-shot adversarial robustness. The authors proposed a text-guided contrastive adversarial training (TeCoA) to be used along with finetuning to improve the zero-shot adversarial robustness. All these studies try to examine the robustness by either formulating transformer-specific attacks, proposing new benchmarks, carefully looking at different architectural components, or optimizing training strategies. However, a proper and common framework can better help compare the various VLPMs. The architectural difference alone makes this a difficult but essential task that needs to be looked at.

\section{Interpretability and Explainability} 
Irrespective of the architecture, it is imperative that we can interpret as well as explain the decisions made by the model. Transformers have relied heavily on attention to provide that explanation. A few methods originally proposed for CNNs have been extended for transformers as well, like GradCAM \cite{selvaraju2017grad}. We have categorized the proposed methods into two categories, namely, gradient and visualization-based methods, and probing tasks. While visualization-based methods usually use inter- and intra-modality interactions to visually explain the decisions, probing tasks are specifically designed to explain a particular aspect or component of the models and can be restrictive. Finally, we discuss attention and how reliable it is as an explanation.


\subsection{Gradient-based and Visualization-based Methods} 
Among several explanation methods proposed in the literature, many have been extended to transformer-based models. We first present the different gradient and visualization-based methods that are more in line with transformers and VLPMs. Attention maps are a well-known method for interpreting transformer models. Modifications of these methods have been proposed in the literature, like the Attention Rollout \cite{abnar2020quantifying}, which combined layers to get averaged attention. \cite{voita-etal-2019-analyzing} modified the LRP method specifically for transformers overcoming the computational barriers. Further, Relevancy Map or HilaCAM \cite{chefer2021generic} uses the self-attention and co-attention modules considering classification tokens appended during downstream tasks and associated values to generate a relevancy map tracking interactions between different modalities and backpropagating relevancies. The method applies to both unimodal and multimodal models. Apart from these methods, VL-InterpreT \cite{aflalo2022vl} is more like a tool that gives an interactive interface looking at interactions between modalities from a bottom-up perspective. It uses four modality attention heads: language-to-vision attention, vision-to-language attention, language-to-language attention, and vision-to-vision attention, allowing it to look at interactions within and between modalities. MULTIVIZ \cite{liang2022multiviz} is another method to analyze multimodal models interpreting unimodal interactions, cross-modal interactions, multi-modal representations, and multimodal prediction. gScoreCAM \cite{chen2022gscorecam} studied the CLIP \cite{radford2021learning} model specifically to understand large multimodal models. Using gScoreCAM, objects can be visualized as seen by the model by linearly combining the highest gradients as attention.

\cite{DBLP:journals/corr/abs-2106-12620} proposes interpretability-aware redundancy reduction ($IA-RED^2$) to make transformer cost-efficient while using human-understandable architecture. The study \cite{DBLP:conf/nips/CheferSW22} manipulates the relevancy maps to alleviate the model’s robustness. Lower relevance is assigned to the background pixels, so the foreground is considered with more confidence. \cite{DBLP:conf/nips/QiangPLLJZ22} proposes the AttCAT explanation method that uses attentive class activation tokens built on encoded features, gradients, and attention weights to provide the explanation. B-cos transformers are proposed by \cite{DBLP:journals/corr/abs-2301-08669}, which are highly interpretable, providing holistic explanations. \cite{DBLP:conf/cvpr/NalmpantisPGPA23} proposes another interpretation method called Vision DiffMask, which identifies the relevant input part for final prediction using a gating mechanism. A faithfulness test is also used to showcase the validity of this post-hoc method, concluding that there is a lack of faithfulness tests in the literature. \cite{DBLP:conf/cvpr/ChoiJH23} proposes Adversarial Normalization: I can Visualize Everything (ICE) to visualize the transformer architecture effectively. It uses adversarial normalization and patch-wise classification for each token, separating background and foreground pixels. The most common theme in these methods is exploiting attention weights and gradients to make the information flow more targeted. Another theme is to extend available metrics by making them computationally effective.

\subsection{Probing Tasks} 
Most of the explanation methods for VLPMs are based on probing tasks. These tasks are designed to study a particular aspect of the model and thus are hard to generalize. VALUE or Vision And Language Understanding Evaluation \cite{cao2020behind} method gives several probing tasks to understand how pre-training helps the learned representations. The authors made several important observations: (i) the pre-trained models attend to language more than vision, something that has been corroborated throughout the literature; (ii) there is a set of attention heads that capture cross-modal interactions; and (iii) plotting attention can depict interpretable visual relations as was corroborated in the previous section as well, among others. \cite{dahlgren-lindstrom-etal-2020-probing} further proposes three probing tasks for visual-semantic space, which are relevant for image-caption pairs and train separate classifiers for probing. The tasks are (i) a direct probing task designed for the number of objects, (ii) a direct probing task for object categories, and (iii) a task for semantic congruence. \cite{hendricks2021probing} furthermore proposes probing tasks for verb understanding by collecting image-sentence pairs with 421 verbs commonly found in the Conceptual Captions dataset~\cite{sharma2018conceptual}. \cite{DBLP:conf/aaai/SalinFAF22} proposed a set of probing tasks to better understand the representations generated by vision-language models, comparing the representations at pre-trained and finetuned levels. Further, datasets are designed carefully for multimodal probing, trying to reduce dependency on bias while making predictions. While probing tasks are helpful and can answer meaningfully with regard to particular problems, they have to be carefully crafted for relevant results and are very specific. At times, extra models or classifiers are required for probing, making the probing tasks applicable to selected models only.

\subsection{Dissecting Attention} \label{dis}
As can be seen in this section so far, attention is heavily used in the methods proposed to explain and interpret VLPMs. In fact, attention is one of the main reasons why transformers have been attributed to working so well. However, recently, attention has been pointed out not to be a reliable parameter for explaining a model’s decision in some studies. For VLPMs, in particular, fusing the modalities can make it difficult to interpret how the attention is distributed and how it should be explained. \cite{serrano-smith-2019-attention} evaluated attention for text classification, concluding that while attention can be helpful with intermediate components, it is not a good indicator for a justification. Further, \cite{jain-wallace-2019-attention} studied the relationship between attention weights and the final decision for several NLP tasks and concluded that attention weights often do not relate to gradient-based methods for computing feature importance; hence, they do not provide helpful or meaningful explanations. 

While these methods concluded that attention is not reliable as a justification tool, the studies have been limited to language-based tasks and need a proper in-depth analysis given how heavily current methods rely on the mechanism to interpret the models. \cite{park2022explanation} computed a relation between the attention map and input-attribution method by proposing Input-Attribution and Attention Score Vector (IAV). It tried to combine attention with attribution-based methods to utilize both components as a justification tool. Such methods can help alleviate this mistrust of attention. \cite{ DBLP:conf/icml/SahinerEOPMP22} studies attention under convex duality that can help provide interpretability for the architecture. \cite{DBLP:conf/icml/LiuLGKL022} takes polarity into consideration along with attention. The authors propose a faithfulness violation test that can help quantify the quality of different explanation methods. We believe that attention needs to be evaluated as an interpretability metric for more vision and vision-language tasks. Combining the module with other established methods, like attribution-based methods, or examining the methods on controlled benchmarks can help.

\section{Open Challenges and Opportunities} 
The previous sections discuss several methods and techniques to make VLPMs fair, robust, explainable, and interpretable. However, they also highlighted a lack of specific architecture-based methods and standard protocols. Even with all the progress, there are several open challenges that require further development and analysis. Here, we discuss some of the open challenges for improving different aspects of the trustworthiness of VLPMs.

\noindent\textbf{Trustworthiness of VLPMs:} The concept of trustworthiness as a whole is lacking in the current analysis of VLPMs. 
A formalized and standardized framework can help set the baselines for the growing number of transformer architectures. One basic need is to make these models just as trustworthy to ensure that their decisions can be trusted and relied upon while staying away from harmful biases like using faithfulness tests for quantifying the model’s explainability. As we continue to use these models for security-critical applications, we need to be able to depend on the models and their decisions.  

\noindent\textbf{Examining Attention:} Attention mechanisms are often used to explain how models make decisions by creating visual representations that provide reasoning behind these decisions. However, to better understand and interpret attention, especially in the context of vision and cross-modality, we need to thoroughly examine attention modules. Analyzing models under adversarial conditions can also help us gain valuable insights and improve our understanding of attention mechanisms. Additionally, attention is a critical factor in ensuring the trustworthiness of transformer models. Therefore, we should examine attention from three different angles: its impact on model performance, its role in explaining decisions, and its role in understanding the model's reasoning.


\noindent\textbf{Probing the Vision Modality:} The literature has time and again iterated that for VLPMs, decisions have a stronger influence from the language modality than the visual modality. We believe a big gap exists between a systematic review of how the vision modality affects decisions and how we can better utilize it to avoid language bias. While tasks like VQA have recognized language bias, VLPM as a generalized architecture has not been explored for this bias as extensively. Better pre-trained tasks aligning the vision modality along with cross-modality interactions can be a way forward for improving the generalization as well as the effect of the vision modality on the entire model. Moreover, vision plays a crucial role in understanding object semantics on tasks like object detection and semantic segmentation, and thus, their reduced influence in vision-language tasks can be seen as a disadvantage. Studying the alignment between vision and text modality can also be a way forward. 

\noindent\textbf{Better Generalized Methods:} There is a need for better generalized methods that can evaluate not only between CNNs and transformers but also between different architecture formats within transformers. Also, with increasing hybrid architectures, such methods can help create a better comparison framework, providing effective baselines for future studies. Some studies \cite{DBLP:journals/corr/abs-2205-09256,DBLP:conf/cvpr/TangWK0LDWLX23} have used one modality to guide the other while training or used one modality to train the multimodal models, which can allow correcting for bias or adversarial vulnerabilities. 

\noindent\textbf{Cross-modality and Universal Weights:} Transformer models are known for their similar architecture, even when processing different modalities. However, the pre-trained weights are not as easily adapted between the modalities, and alignment remains an open challenge. Aligning the two modalities can help improve the representations for VLPMs and better project the two modalities in a similar projection space. A universal model that can represent both modalities similarly can help with performance as well as robustness, however, there is still a gap in getting universal pre-trained weights that can adapt to different modalities and require further research. 

\noindent\textbf{Strategic Pre-training:} Pre-training has been demonstrated to be beneficial for transformers, but it is costly. It can be a tedious process that requires large datasets and pre-training tasks that utilize heavy computing power. We have also seen how these large datasets can be a potential source of bias. With better and more focused pre-training strategies \cite{DBLP:conf/aaai/ZhouPZHCG20}, the training cost can be reduced while improving task-aware performance. With proper strategies in place, bias at the pre-training stage can be mitigated or avoided during finetuning.

\noindent\textbf{Interplay of VLPMs with Audio Models:} In several multimedia applications ranging from audio-visual scene comprehension to speech-driven image recognition and immersive human-computer interactions, the fusion of vision, language, and audio plays a pivotal role. Consequently, it becomes imperative to explore the interplay between audio models and VLPMs to enhance our capabilities in perception, understanding, and communication, thereby offering more enriched and immersive experiences.


\noindent\textbf{Responsible ML Datasets:} The trustworthiness of VLPMs and transformer models is intricately tied to their training data. These algorithms learn patterns from the data they are exposed to, which may inadvertently incorporate any inherent flaws present in the data, thereby influencing their behavior. Therefore, it is important to understand the crucial role of Responsible Machine Learning Datasets \cite{mittal_RMLD}, encompassing aspects such as privacy \cite{saheb_ijcai} and adherence to regulatory standards. In addition, \textit{machine unlearning} concepts should be explored to ensure these systems can adapt and comply with evolving regulatory norms.

\section{Discussion} 

Despite the remarkable human-like performance demonstrated by Vision-Language Pre-trained Models (VLPMs) and Vision Transformers, it is of paramount importance not to underestimate the crucial dimension of trustworthiness. As VLPMs continue to gain widespread adoption on a global scale, a rigorous examination becomes imperative. This paper presents a comprehensive analysis of VLPMs, addressing three essential dimensions: bias/fairness, robustness, and explainability/interpretability. Firstly, we scrutinize biases within VLPMs, recognizing that while datasets often serve as the primary source of bias, biases can also seep into the models and algorithms themselves. Addressing this issue requires a thorough evaluation and mitigation study, a challenge further complicated by VLPMs' multidimensional nature encompassing both vision and language. Establishing a robust framework is essential to conduct bias assessments tailored to these complex models effectively. Next, we discuss about the robustness of VLPMs. While VLPMs have been extensively compared to their CNN counterparts, a noticeable gap exists when it comes to architecture-specific studies that explore vulnerabilities unique to VLPMs. Finally, we explore VLPMs using visualization-based and probing methods, which, although limited in availability, provide valuable insights to enhance our comprehension of VLPMs' inner workings. We also highlighted some of the open challenges confronting VLPMs. We hope that this study serves as a foundation for researchers to identify gaps and work towards enhancing both the performance and trustworthiness of these models.

\section{Acknowledgement}
The work is partially supported through the grant from Technology Innovation Hub (TIH) at IIT Jodhpur. M. Vatsa is partially supported through the Swarnajayanti Fellowship. 

\bibliography{aaai24}

\end{document}